\documentclass[sn-mathphys-num]{sn-jnl}

\usepackage{graphicx}%
\usepackage{multirow}%
\usepackage{amsmath,amssymb,amsfonts}%
\usepackage{amsthm}%
\usepackage{breqn}
\usepackage{array}
\usepackage{mathrsfs}%
\usepackage[title]{appendix}%
\usepackage{textcomp}%
\usepackage{manyfoot}%
\usepackage{booktabs}%
\usepackage{algorithm}%
\usepackage{algorithmicx}%
\usepackage{algpseudocode}%
\usepackage{listings}%
\usepackage[dvipsnames]{xcolor}

\theoremstyle{thmstyleone}%
\theoremstyle{thmstyletwo}%

\theoremstyle{thmstylethree}%

\begin{document}

\title[Article Title]{$D^3$ - Deep Deconvolution Deblurring for Natural Images }


\author*[1]{\fnm{Vamsidhar } \sur{Saraswathula}}\email{sreeramvds@gmail.com}

\author[2]{\fnm{Rama Krishna} \sur{Gorthi}}\email{rkg@iittp.ac.in}


\affil[1,2]{\orgdiv{Electrical Engineering department}, \orgname{Indian Institute of Technology (IIT) Tirupati}, \orgaddress{\street{}, \city{Tirupati}, \postcode{517619}, \state{}, \country{India}}}




\abstract{ In this paper, we propose to reformulate the blind image deblurring task to directly learn an inverse of the degradation model represented by a deep linear network. We introduce ``Deep Identity Learning (DIL)”, a novel learning strategy that includes a dedicated regularization term based on the properties of linear systems, to exploit the identity relation between the degradation and inverse degradation models. The salient aspect of our proposed framework is it neither relies on a deblurring dataset nor a single input blurry image (e.g. Polyblur, a self-supervised method). This framework detours the typical degradation kernel estimation step involved in most of the existing blind deblurring solutions by the proposition of our Random Kernel Gallery (RKG) dataset. The proposed approach extends our previous Image Super-Resolution (ISR) work, NSSR-DIL, to the image deblurring task. In this work, we updated the regularization term in DIL based on Fourier transform properties of the identity relation, to deliver robust performance across a wide range of degradations. Besides the regularization term, we provide an explicit and compact representation of the learned deep linear network in a matrix form, called ``Deep Restoration Kernel (DRK)" to perform image restoration. Our experiments show that the proposed method outperforms both traditional and deep learning based deblurring methods, with at least an order of 100 lesser computational resources. The $D^3$ model, both L-CNN \& DRK, can be effortlessly extended to the Image Super-Resolution (ISR) task as well to restore the low-resolution images with fine details. The $D^3$ model and its kernel form representation (DRK) are lightweight yet robust and restore the blurry input in a fraction of a second.}




\maketitle

\section{Introduction}
\label{intro}
Image deblurring is a well-established low-level vision task that aims to generate a sharp image from the given corresponding blurry input observation(s). In image formation, the blur degradation can be caused by various factors like, but not limited to, camera shake, observed relative motion between camera and target, camera defocus, camera lens aberration, light diffraction due to finite lens aperture, integration of light in the sensor \cite{polyblur}. Image deblurring is a severely ill-posed problem as multiple combinations of a sharp image and a blur kernel can result in the same blurry image. However, practical applications in prominent domains like medical imaging, satellite imaging, and surveillance demand a sharp version of the scene of interest for its analysis.
\par Image deblurring solutions can be broadly categorized into non-blind and blind approaches. Non-blind approaches aim to restore clear images from blurry inputs when the degradation kernel is known \cite{Wiener, richardson1972bayesian}. In contrast, blind deblurring approaches work with an unknown degradation kernel. The blind deblurring works either estimate both the blur kernel and the sharp image simultaneously \cite{mirani, nsm, cai2009blind, perrone2014total} or the blur kernel is estimated first and then non-blind deblurring is performed \cite{fergus2006removing, cho2009fast, pan2017deblurring}.
The traditional methods to address the image deblurring task using either non-blind or blind approaches include Maximum a-Posterior (MAP) approaches with a variety of data terms \cite{data1, data2} and sophisticated prior terms \cite{prior1, prior2, sprior1, sprior2}, Variational inference methods \cite{var1, var2}, Regularization techniques \cite{nsm,NLtknv} and Sparse representation methods \cite{cai2009blind, sparsereg1}.
 It is demonstrated that estimating the blur kernel is easier than jointly estimating the blur kernel and sharp image together \cite{understand}. Later, since 2014, remarkable improvement in the restored sharp image quality has been observed with the advent and continuous advancements of Deep Learning (DL) techniques. The recent deblurring solutions, including the pioneering works \cite{sun2015learning, nah2017deep}, adopt sophisticated techniques like Residual blocks \cite{purohit2020region, tao2018scale, dun}, Generative adversarial networks \cite{kupyn2018deblurgan,zhang2020deblurring}, Recurrent Neural Networks \cite{zhang2018dynamic, park2020multi}, Transformers \cite{liang2021swinir, zamir2022restormer}, and specialized loss functions \cite{loss1, loss2}.
\par Despite vigorous research efforts, the existing deblurring algorithms still exhibit the inverse correlation between accuracy and computational efficiency. Hence, these methods are not very suitable for practical applications. Regardless of their computational efficiency the traditional methods (e.g.: \cite{mirani, partialdeconv, richardson1972bayesian, sprior1}) fail to handle a wide range of degradations and cast the deblurring problem into an iterative framework, which is computationally expensive. Additionally, the optimization problems of these lucid models limit their practical usage. In contrast, the State-of-the-Art (SotA) DL based methods are promising but are entirely data-driven and incur significant expenses in terms of time and computational resources for training and implementation. Furthermore, DL models grapple with generalization issues and encounter challenges when operating within memory-constrained environments due to their sizable footprint. Further, the deblurring performance of the model is challenged by the error-free assumption on the blur kernel. In practice, the estimated blur kernel is prone to error. Even the slight deviation in the estimated blur kernel leads to the ill-posedness of the problem as it results in a great mismatch of the deconvolution kernel followed by the notable artifacts in the deblurred image \cite{ji2011robust, li2023self}. The extensive dependency on datasets for prior information could limit the generalization ability of blind approaches. \\
For practical applications, few such solutions exist that deliver reliable deblurring performance while maintaining accuracy, speed, and computational efficiency \cite{polyblur, 1shot}. These methods are self-supervised and offer image-specific solutions. These methods compute the deconvolution kernel and apply it directly to the given input blurry image. However, the degradation kernel estimation step involved in \cite{polyblur, 1shot} is computationally expensive and required for every input. Besides, these methods await the blurry input to estimate the blur kernel and its deconvolution inverse.
\par In this paper, we redefine the task of learning the deblurring model from image data to merely computing an inverse of the degradation kernel space. Here, the degradation model is represented with a wide set of anisotropic Gaussian kernels perturbed with noise called as `Random Kernel Gallery (RKG)' dataset and thus detours the error-prone blur kernel estimation step. We adopt the generalized Gaussian distribution to construct the RKG dataset since many point spread function (static blur) applications are symmetric and can be modeled by such distribution \cite{1shot, polyblur, partialdeconv}. We introduce a novel learning strategy called “Deep Identity Learning” (DIL) that exploits the identity relation between the degradation model and its convolution inverse. This method builds on our previous work ``NSSR-DIL: Null Shot Image Super-Resolution Using Deep Identity Learning", a computationally efficient Image Super-Resolution (ISR) model. Here, in this work, we made the DIL objective function robust with the updated regularization term based on the Fourier transform properties of the identity system and spatial domain properties of the convolution operator to effectively ensure the identity system properties of our inverse degradation model. We characterize the deconvolution model using a custom lightweight Linear Convolutional Neural Network (L-CNN) to train on our DIL objective using the RKG dataset. Unlike existing blind self-supervised deblur frameworks like \cite{polyblur, 1shot}, the proposed deblurring framework is image-independent and doesn't require blurry or pair of blurry and sharp image data at any stage of the deblur task learning. Furthermore, due to the non-linear and deep nature of the existing deblur works in the literature, providing the explainability of their inverse degradation operation was not feasible or trivial. Therefore, besides the modified regularization term, we provide a compact and effective representation of the proposed L-CNN, in a matrix form, called ''Deep Restoration Kernel (DRK)" to directly apply and deconvolve the blurry input. Unlike the traditional methods that estimate deconvolution kernel for a single degradation kernel for the blurry image assuming the degradation in a specific way, we estimate the deconvolution kernel (i.e. DRK) for the degradation kernel space. Here, we consider many possible ways that an image could be blurred, which allows for a more accurate and effective way to restore the image. The effective representation of the deep model L-CNN in a matrix form i.e. DRK, provides an understanding of the learning mechanism of that deep model (more details are provided in section \ref{sec:drk}). The proposed work can be considered as the ``model-aided deep learning framework” for the deblur task unlike the current SotA works that do “distribution mapping from degraded to clean image data” using DL frameworks. In our experiments, our proposed computationally efficient L-CNN and its effective representation DRK were found to perform the deblur tasks across a wide range of degradations more effectively. The highlights of our proposition are as follows. 
\begin{enumerate}
    \item To the best of our knowledge, $D^3$ is the first DL-based \textbf{image data-independent} deblurring framework. 
    \item We develop the deblurring model without blurry input image and its corresponding degradation prior.
    \item We propose a novel learning function i.e. DIL based on identity system properties with an updated regularization term driven by the spatial and frequency domain inferences to effectively learn an inverse degradation model.   
    \item We provide an effective and explainable, explicit representation in matrix form, of the learned inverse degradation model (L-CNN), i.e. DRK.
    \item The proposed $D^3$ model and the DRK perform superior deblurring, against the existing SotA deblur methods while being at least 100 times more computationally efficient. 
    \item We do demonstrate that the proposed $D^3$ framework can be readily extended to the Image Super-Resolution (ISR) task.
\end{enumerate} 
The overall organization of the rest of the paper is as follows. Section 2 presents Related work. A detailed explanation of our proposed $D^3$ method is given in Section 3. In Section 4, a thorough experimental setup and results are presented, followed by a Discussion in Section 5 and a Conclusion in Section 6.
\begin{figure}[t]
    \centering
    \includegraphics[width = 0.9\textwidth, height = 3.75cm]{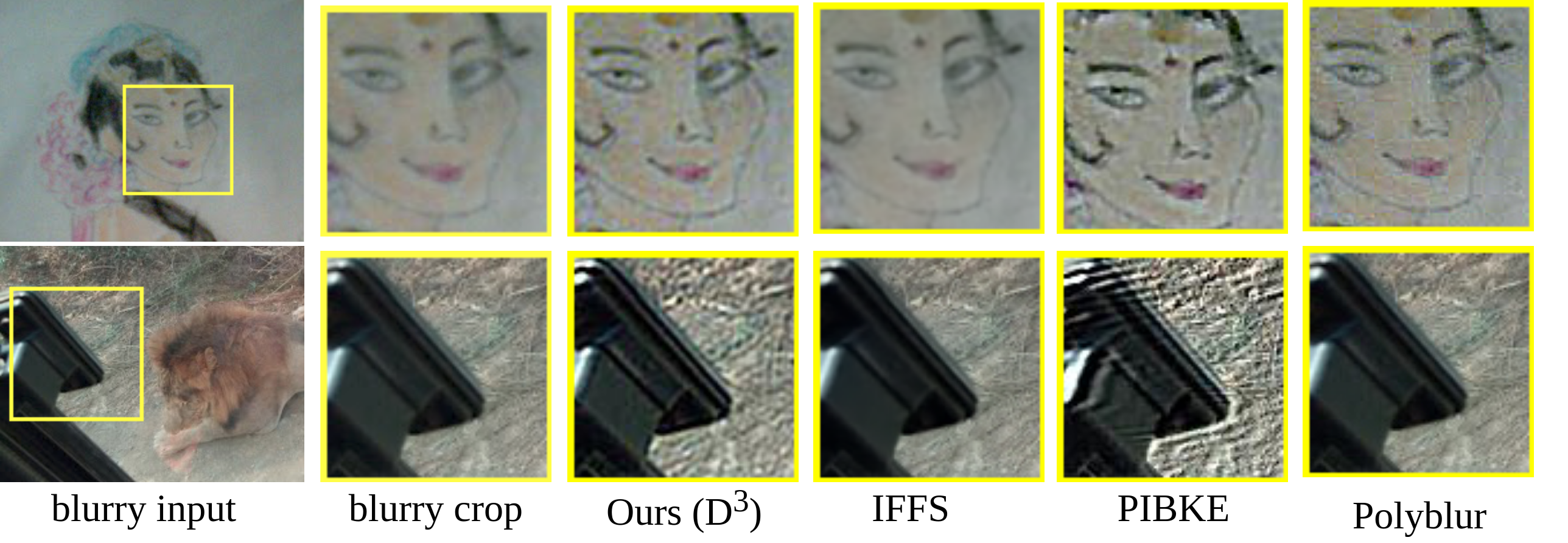}
    \caption{Qualitative results on real blurry inputs captured using a Samsung J7 prime smartphone camera (13MP, f/1.9). }
    \label{fig:realresult}
\end{figure}

\section{Related work}
\label{relatedwork}
This section presents an overview of the existing related deblurring works.\\
\par \textit{Image sharpening} methods are simple and intuitive to generate sharp images. These methods share a common framework, the input image is filtered and then a proportion of the residual image is added back to the original input. This procedure boosts high-frequencies and increases the contrast in the generated image. The most well-established works include Unsharp Masking and its adaptations. These are simple and effective but sensitive to noise and generate artifacts \cite{polesel2000image, russo2007image, kim2005optimal}. The bilateral filter \cite{tomasi1998bilateral} and its variants \cite{he2012guided, buades2005non} were used to enhance the local contrast and high-frequency details in an image assuming the filtered output is a local transform of the guidance image.  Here, the guidance image can be the same input blurry or any other similar texture image that is
not heavily degraded. Based on local structure and, sharpness information \cite{zhu2011restoration} proposed an adaptive sharpening method that performs simultaneous noise reduction and sharpening. However, the deblurring performance of these methods is limited since they do not explicitly estimate image blur. These methods are strictly local operators.
\par \textit{Gaussian deblurring}, as one of the causes of out-of-focus is usually modeled as a Gaussian function \cite{liu2020estimating}. Most of the Gaussian deblurring methods \cite{honarvar2014image, carasso2003apex, xue2014novel}, estimate the blur kernel (i.e. Gaussian) parameters like variance, location of mean, and its orientation and then apply non-blind techniques to perform the deblurring task. However, these methods are sensitive to noise and require high inference time. Also, the error in the estimates of the Gaussian kernel parameters leads to over-sharp or over-smoothed results with artifacts \cite{efrat2013accurate}.
\par \textit{Self-supervised learning} works \cite{li2023self, ren2020neural, li2022supervised, chen2022self}, achieves uniform and nonuniform deblurring tasks from the given single test input blurry image. In \cite{ren2020neural,li2022supervised}, two neural networks were used to predict the sharp image and the blur kernel using a total-variation (TV) based MAP estimator. Bayesian inference implemented via Monte-Carlo sampling and \cite{chen2022self} using deep neural network-based ensemble learning technique respectively for uniform deblurring. \cite{li2023self} addresses both uniform and nonuniform motion deblurring using the Monte Carlo expectation maximization algorithm. These methods require high-inference time and are computationally complex though. Polyblur \cite{polyblur} is a blind and self-supervised approach that uses polynomial functions to restore image details blurred by mild camera shake or motion blur supporting real-time applications. It estimates the blur kernel of the image by analyzing the distribution of the gradient in the image. Then it applies the kernel multiple times, based on the severity of the blur. The work 1Shot-MaxPol \cite{1shot}, estimates the degradation kernel using scale-space analysis in the frequency domain of the input blurry image. Then finds the deconvolution kernel, as a linear combination of finite impulse response derivative filters, that can directly convolve with the blurry input using a closed-form solution.\\
Towards addressing the limitations of data dependency, time, and computational complexities for the practical use case of real-time implementation, we proposed a blind lightweight yet robust purely image data-independent model to restore the blurry input image.
\section{Proposed method}
\label{proposed}
In this section, the introduced RKG dataset, the proposed $D^3$ model, dedicated Regularization, and explainable DRK are discussed in greater detail.
\subsection{Problem formulation}
\label{problem}
In general, the End-to-End Deblurring (E2ED) framework can be observed as the degradation model followed by its inverse degradation model. Both the input and output of the E2ED model are sharp images. At first, the degradation model outputs a degraded version of the sharp image i.e. the blurry image, depicting the degradation process. Later, the inverse degradation model restores the sharp image from the obtained blur image i.e. the sharp image, through the restoration process. The degradation and inverse degradation models are discussed below.\\
\begin{figure*}[t]
\centering
\includegraphics[width=\textwidth ]{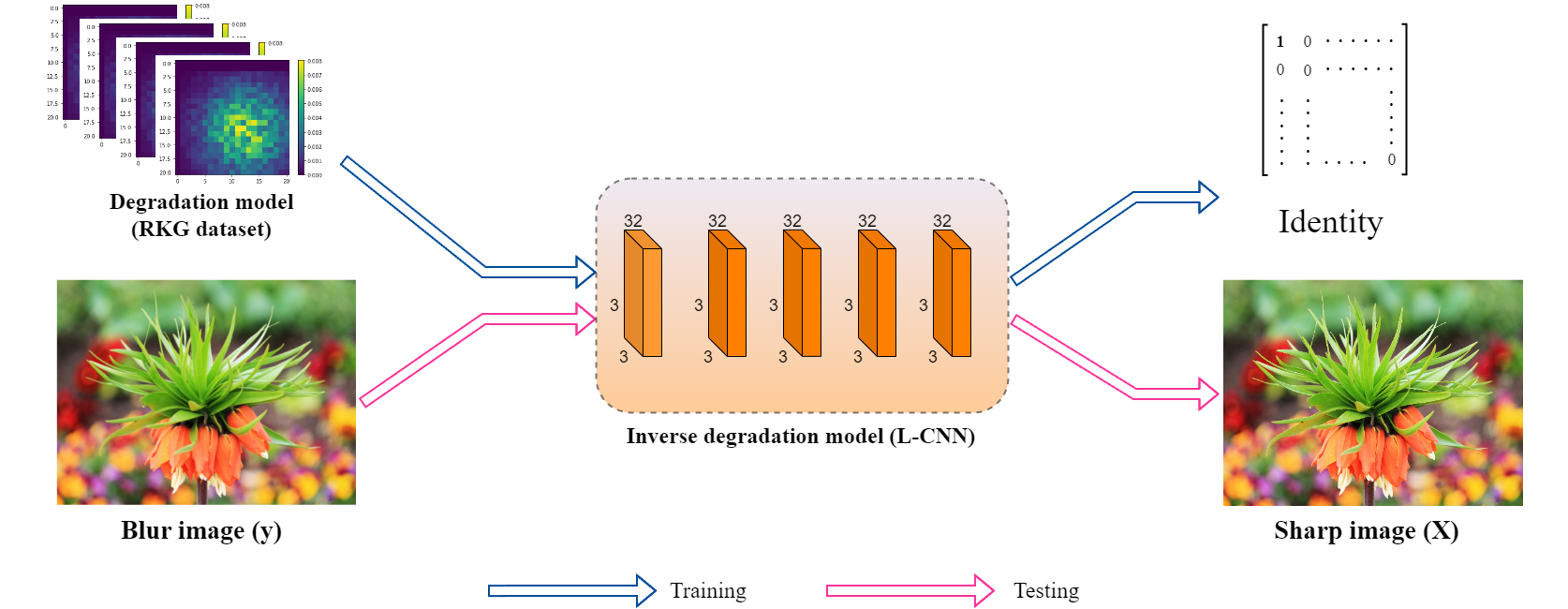}
\caption{{The training and inference methodology of the proposed $D^3$ model.}}
\label{fig:$D^3$}
\end{figure*}
\textcolor{RubineRed}{\textbf{Degradation model}:} 
The degradation model typically represents the convolution operation between the input sharp image and the degradation kernel ($K$), which results in a degraded image i.e. blurry image. The mathematical representation of the correspondence between sharp image $X \in {R}^{M \times N}$ and blur image $y \in {R}^{M \times N}$ is given in Eq. (\ref{eq1}), assuming the spatial invariant blur \cite{understand}. 
\begin{equation}
    \label{eq1}
    y = X \ast K 
\end{equation}
Here $\ast$ represents the convolution operation.\\
\textcolor{RubineRed}{\textbf{Inverse degradation model}:}
The objective of the deblurring task is to achieve the inverse of the degradation model i.e. to reconstruct the sharp image from the given blurry image input. The mathematical representation of the inverse degradation model is given in Eq. (\ref{eq2}).  
\begin{equation}
    \label{eq2}
 y \ast K^{-1} = X
\end{equation}
Here $K^{-1}$ represents the inverse degradation kernel.\\
\textcolor{RubineRed}{\textbf{End-to-End Deblurring (E2ED) model}:} The unified degradation and inverse degradation models form an E2ED problem. Here, it is important to observe that the unified framework represents an ``\emph{identity model}'' with the same entity as the input and also as the output i.e. sharp image (in the ideal case). Here the degradation and inverse degradation models are characterized and effectively represented by their respective kernels $K$ and $K^{-1}$. Therefore, for an ideal deblurring, the convolution operation between $K$ and $K^{-1}$ should result in an identity relation. In practice, the output of the inverse model is the restored image with high-frequency information from the blurry input image.\\
Based on the observed E2ED model, we propose a simple, computationally efficient, and novel framework that is independent of images to learn the deblurring task. Here, we propose to learn the inverse degradation model forming an identity relation with the degradation kernel. The identity relation is given in Eq. (\ref{eq3}). The proposed formulation simplifies the task of learning the deblurring problem from image datasets using deep architectures with several millions of learnable parameters, to the task of \textit{ ``identity'' learning} between the degradation kernel $K$ and the inverse degradation kernel $K^{-1}$ through a custom linear CNN (discussed in Sec.\ref{L-CNN}). Hence, we term the proposed identity learning using a deep model as ``\emph{Deep Identity Learning (DIL)}". 
\begin{equation}
    \label{eq3}
 K \ast K^{-1} =  ~\delta
\end{equation}
Where $\delta$ is a two-dimensional impulse function of dimension $P \times Q$, and $P, Q$ values are based on the dimensions of $K$ \& $K^{-1}$.
\subsection{Random Kernel Gallery (RKG) Dataset}
\label{kdataset}
To realize the identity learning task, a set of degradation maps sampled from the degradation kernel space is generated. In deblurring literature, the degradation modeling is mainly carried out conventionally by using Gaussian prior \cite{polyblur,liu2020estimating}. Theoretically, $K$ is modeled as unimodal \cite{npbsr} \&  Gaussian \cite{CRMnBISR}. The existing well-known blind Deblurring works like \cite{polyblur} consider the $K$ to be isotropic or anisotropic Gaussian models. Also, the work \cite{understand} demonstrated that with an appropriate estimation algorithm, blind deconvolution can be performed even with a weak Gaussian prior. Following these widely adopted premises, we constructed a set of degradation models with the anisotropic Gaussian kernels.\\
In this work, a set of anisotropic Gaussian kernels with dimensions $11 \times 11$, having $\sigma_1, \sigma_2 \in U[0.175, 3]$ and rotation angle $\theta \in U[0, \pi]$ were randomly generated as representative samples of the degradation kernel space. The number of $K$ samples required to effectively learn the inverse degradation kernel $K^{-1}$ is determined empirically as $2,400$. The reference output to learn the inverse degradation model is the two-dimensional impulse function mentioned in Eq.\ref{eq3}. We term this dataset as the ``Random Kernel Gallery (RKG)'' dataset.\\
The proposed $D^3$ method is not limited to the ``RKG-dataset'' with anisotropic Gaussian kernels. $D^3$ can be re-trained on any other kernel distribution if proven to be a more accurate representation of the degradations kernel space.
\subsection{Linear Convolutional Neural Network (L-CNN)}
\label{L-CNN}
Deblurring is an ill-posed inverse problem for which a unique inverse will not exist. Computing a single-layer network to learn the inverse of the degradation kernel cannot serve the deblurring task's objective. This is because a matrix/single layer accepts only one set of parameters/global minima with convex loss. Also, the $K$ can usually be a low-rank matrix. Further, it was empirically found that single-layer architecture does not converge to the correct solution \cite{choromanska2015loss}. Whereas the multi-layered linear networks have many good and equally valued local minima. This allows many valid optimal solutions to the optimization objective in the form of different factorizations of the same matrix \cite{kawaguchi2016deep}, \cite{saxe2014a}, \cite{arora2018optimization}. Following these research results, we propose a multi-layer Linear CNN (L-CNN) with no activations to learn the inverse degradation kernel, with the degradation kernels ($K$) from the RKG dataset as its input. The proposed L-CNN is a computationally efficient architecture having a depth of five layers and a width of 32 with $3 \times 3$ filters across the depth. Here the L-CNN is chosen to maintain the dimensions of the input throughout the network i.e. the same output dimension at every layer. Therefore, at the inference stage, L-CNN operates on blur images to generate sharp images with fine details. \\
The general limitations of the networks that maintain the dimensions of the input at the output, like computational time, complexity, and memory are because of operations in high dimensional space. It is important to note that these limitations are not valid in this work as the input to the CNN is kernel $K$, a matrix of very small dimension (i.e. $11 \times 11$), compared to very high dimensional or 3-D tensor input images, in practice. The learning and inference methodology of the $D^3$ model is depicted in Fig. \ref{fig:$D^3$}.
\subsection{Deep Identity Learning (DIL)} The L-CNN is trained on the RKG dataset with the DIL objective given in Eq. (\ref{eq4}).
\begin{equation}
    \label{eq4}
  Loss(L) = ~ \lVert K \ast K^{-1} - ~\delta \rVert^2_2 + R 
\end{equation}
Here, $\lVert K \ast K^{-1} - ~\delta \rVert^2_2$ term ensures the identity relation between the degradation and its inverse model and $R$ is the proposed regularization, defined in Eq. (\ref{eq5}). 
\begin{equation}
\label{eq5}
  R = \lambda_1 \times R_1 +\lambda_2 \times R_2  +\lambda_3 \times R_3
\end{equation}
where,
\begin{enumerate}
    \item $R_1 = L_{ConvArea} = ~\mid 1 - \sum_{i,j}K^{-1}_{i,j}\mid$; This term ensures the validity of the area property of the convolution operation. It states that the product of the area under the two input signals is equal to the area under their convolution output signal. Consider the identity relation given in Eq. \ref{eq3}, the input signals are degradation kernel and its inverse i.e. $K$ and $K^{-1}$, while the output signal is a 2-D impulse function. The area property of the convolution operation is discussed in detail below.\\   
        The 2-dimensional Fourier transform can be written as
       \begin{equation}
       \label{eq9}
       \mathcal{F}(K(m,n)) = \hat{K}(p,q) = \sum_{m}^{}\sum_{n}^{} K(m,n) \exp^{-j({\frac{2\pi pm}{N_1}}+{\frac{2\pi qn}{N_2}})}
       \end{equation}
       Here, $\mathcal{F}(.)$ represent the Fourier transform operation, (m,n) and (p,q) represent the index locations of the degradation kernel matrix in the Time-domain and Frequency-domain respectively, $\hat{K}(.)$ represent the Fourier transform of $K$  and $N_1, N_2$ represent the dimensions of degradation kernel.
       \\
     We apply the Fourier transform to Eq. \ref{eq3}. ~$\therefore ~\mathcal{F}(K * K^{-1}) = \mathcal{F}(\delta)$;
       \\ Since Convolution operation in the Time domain is equivalent to Multiplication in the Frequency domain (and vice-versa), the above expression can be written as $\hat{K}(p,q). \hat{K^{-1}}(p,q) = 1$~ ($\because ~\mathcal{F}(\delta) = 1$). Here '.' represents the element-wise multiplication operation.      
       \\
        Keeping the values of $(p,q) = (0,0)$ in the Eq. \ref{eq9}, we arrive at another important property of the convolution operation that states the area under the Time-domain signal is equal to the value at the origin in the Frequency-domain form of that signal. Therefore, $\hat{K}(0, 0) = \sum_{m}^{}\sum_{n}^{} K(m,n)$.\\
       The $\sum_{m}^{}\sum_{n}^{} K(m,n)$ is set to be 1 in our RKG dataset, $K$ being a blur kernel. To satisfy the relation  $\hat{K}(0,0).\hat{K^{-1}}(0,0) = 1$, the area under the inverse degradation kernel is imposed to be 1 i.e. $\sum_{m}^{}\sum_{n}^{} K^{-1}(m,n) = 1$ as a regularizer in our proposed DIL objective. 
    \item $R_2 = L_{zerophase} = ~\mid M_0 - ({\angle \mathcal{F}(K) + \angle \mathcal{F}(K^{-1}})) \mid$; enables the sum of phase of degradation kernel $K$ and the inverse degradation kernel $K^{-1}$ to be equal to $0$. Here $M_0$ represents a matrix of zeros having dimensions same as $\delta$ (in Eq.\ref{eq3}).  This term ensures no phase degradation in the overall system. $\lambda_2$ is a hyper-parameter. 
    \item $R_3 = L_{onemag} = ~\mid M_1 - (\mid \mathcal{F}(K) \mid. \mid \mathcal{F}(K^{-1})\mid) \mid$; enables the product of magnitude of the degradation kernel $K$ and the inverse degradation kernel $K^{-1}$ to be equal to 1. Here $M_1$ represents a matrix of ones having dimensions the same as $\delta$ (in Eq.\ref{eq3}). This term ensures no shape degradation of the input image and also ensures the presence of all frequencies in the output same as the input of the E2ED model i.e. sharp image. 
\end{enumerate}
Ideally, the \emph{identity model} (i.e. $K * K^{-1} = \delta$) should behave like an all-pass system with a unit magnitude and with zero-phase so that the input and output of the \emph{identity model} will be identical. Hence the terms $L_{zerophase}$ and $L_{onemag}$ were included in learning the deblurring task. Here, $\lambda_1$, $\lambda_2$, $\lambda_3$ are hyper-parameters. \\ Note that the proposed regularization terms are not general regularization terms just to facilitate the deep learning model to achieve the task. Each term in Eq. \ref{eq5} is driven by the linear systems point-of-view and dedicated to delivering robust deblurring performance, besides constraining the solution space of the inverse degradation kernel for being a valid inverse of the degradation kernel space.\\
\textbf{Note}:  It was observed that the reference label i.e. two-dimensional impulse function, generated using the Python code has unit magnitude but non-zero phase properties. Hence to adjust the phase to be equal to zero and having unit magnitude, the $1$ located at the center is shifted to the very location of the matrix. 
\subsection{Deep Restoration Kernel (DRK)}
\label{sec:drk}
\begin{figure}[!h]
    \centering
    \includegraphics[width = \textwidth, height = 3.5cm]{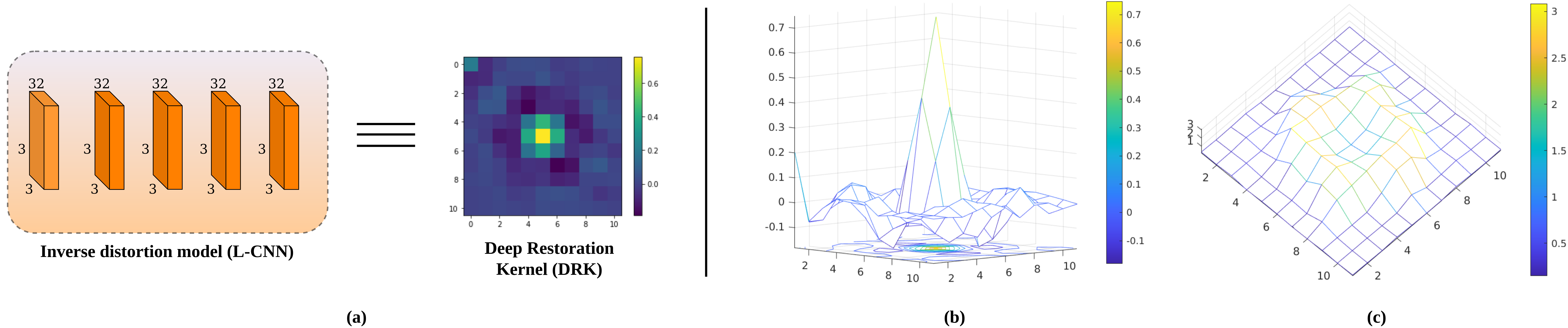}
    \caption{(a) The L-CNN and its effective representation, DRK. The mesh plots of the DRK  (b) spatial domain, (c) The top view of its magnitude spectrum.}
    \label{fig:L-CNN_DRK}
\end{figure}
We proposed L-CNN with solely convolution layers in the interest of extracting the inverse degradation kernel $K^{-1}$, from the trained network. According to the \textit{associative property} of the Linear Time-Invariant (LTI) system, the series interconnection of any arbitrary number of LTI systems is equivalent to a single system. And, the impulse response of the `n' cascaded LTI systems is the convolution of their individual impulse responses. Therefore, an effective representation of the L-CNN model can be obtained in a single matrix form by convolving all the filters of L-CNN sequentially with stride $1$ with impulse as its input. The extracted $K^{-1}$ from the L-CNN is referred to as ``Deep Restoration Kernel (DRK)''. It gives the explicit matrix representation of the inverse degradation model for the deblurring problem. The DRK is equivalent to the deep L-CNN model and can be presented as a filter as shown in Fig. \ref{fig:L-CNN_DRK} (a).\\
\textbf{\textcolor{RubineRed}{Explainability:}} The explicit representation of the inverse degradation kernel i.e. DRK, assists in the progress of understanding the learning mechanism of the deep model (i.e. L-CNN) for the deblurring task. The DRK, a matrix, and the L-CNN are the inverse instances of the degradation space spanned by the RKG dataset. The mesh plots of the DRK were presented in the spatial domain in Fig. \ref{fig:L-CNN_DRK} (b) and the top view of its magnitude spectrum in  Fig. \ref{fig:L-CNN_DRK} (c). It can be vividly observed that the magnitude spectrum of the DRK has the characteristics of both low-pass and high-pass filters. It has a magnitude equal to one around the center (i.e. low-pass filter) and increasing magnitude away from the center (i.e. high-pass filter). The spatial domain form of DRK has a Mexican hat-like structure. Therefore, it can be noted that the extracted DRK can preserve the low frequencies and enhance the high frequencies, acting as a high-frequency-boosting filter or unsharp masking filter. Henceforth, it can be inferred that the proposed deep model can learn the actual deconvolution task, unlike the existing DL-based deblurring models which learn mapping from blurry data to sharp data distribution.

In this way, we can have a deeper understanding of the weights that were obtained from a trained deep model and explain the trained model. Therefore, the proposed $D^3$ model for the deblurring task is \textbf{no longer a black box} deep learning model.\\
We also note that this DRK alone could also be employed for the Deblurring task as per Eq. \ref{eq2} (refer Fig. \ref{fig:L-CNN_DRK}), and the results were presented in Table \ref{tab:comparison}. This makes the proposed method highly efficient for real-world applications as a single matrix with parameters in the order of 100, serves the deblurring task effectively. Also, the computational complexity was reduced from processing the blurry image through a deep model to the convolution operation between the DRK and the blurry image. The effective representation of the L-CNN facilitates the application of the regularization constraints on it easily. With DRK, it is a unique restoration process as it provides ample scope for the development of explainable models for deblurring.
\section{Experiments}
In this section, the implementation details, deblurring results, the effect of the proposed regularization term, and other related studies were discussed.
\subsection{Training setup}
We trained the proposed L-CNN (refer Sec.\ref{L-CNN} for architecture details of L-CNN) on the RKG dataset (refer Sec. \ref{kdataset}) with the learning objective given in Eq. (\ref{eq4}). The number of epochs was $40$ and the learning rate was $10e^{-4}$. The Adam \cite{adam} optimization was used, with $\beta =0.9$. The values of hyper-parameters used in Eq. (\ref{eq5}), set empirically, are as follows, $\lambda_1 = 0.8,~\lambda_2 = 0.8,~\lambda_3 = 0.4$. The $D^3$ code and RKG dataset will be made publicly available soon for reproducibility.  We have used a computer with an NVIDIA-GTX 2080 Ti 11GB GPU in all our experiments.
\subsection{Computational complexity:} The proposed L-CNN is an efficient and lightweight deblurring model. The computational complexity in terms of parameters (for DL based approaches) and inference time of the $D^3$ model was compared to SotA methods and is given in Table \ref{tab:comparison}. Here, the proposed $D^3$ model requires very less computational resources, by order of 1000 in terms of parameters and inference time when compared to the recent SotA work i.e. IFFS \cite{IFFS}. Besides $D^3$, the deblurring task with DRK further reduces the required computational resources parameters and inference time by an order of 10 to $D^3$ and by an order of 10K to SotA methods. Thus the proposed $D^3$ model is more suitable for real-time deblurring tasks.
\subsection{Results}
\label{sec:results}
The deblurring performance of our proposed method is compared with a wide range of methods including standard and recent non-DL blind deblurring methods NSM \cite{nsm}, 1Shot-MaxPol \cite{1shot}, Guided Image Filtering (GIF) \cite{GIF}, PIBKE \cite{PIBKE}, Polyblur \cite{polyblur} and DL-based methods DeblurGAN \cite{kupyn2018deblurgan}, USRNet \cite{dun}, XYDeblur \cite{xy}, IFFS \cite{IFFS}. We note that the proposed $D^3$ is only the DL approach among the above-mentioned DL-based methods that achieve the deblurring task without training on image data. \\
\begin{table}[]
\centering
\caption{
Comparison of deblur results on simulated blur dataset in terms of PSNR$\uparrow$, SSIM$\uparrow$ along with the number of parameters (in Millions) and Inference time (in seconds) of various methods were given below. Here, the \textcolor{red}{red} indicates the best score and the \textcolor{blue}{blue} indicates the second-best score.}
\label{tab:comparison}

\begin{tabular}{llllll}
                                                                  \textbf{Method} & 
                                                                  \textbf{Approach} &
                                                                  \textbf{\begin{tabular}[c]{@{}c@{}} \# Parameters \\ (in Millions)\end{tabular}}&
                                                                  \textbf{\begin{tabular}[c]{@{}c@{}}Inference\\ time\\(in seconds)\end{tabular}}&
                                                                  \textbf{PSNR} & \textbf{SSIM}\\ \hline
NSM                 &Supervised/Non-DL                            &   N.A.    &120            &  \textcolor{blue}{27.51 }  & 0.8170   \\ \hline
PIBKE                    &Self-supervised/Non-DL                &  N.A.            &120               &  24.86     & 0.7384   \\
\hline
GIF &Supervised/Non-DL   & N.A. &60 &19.09  &0.6614\\
\hline
1Shot-MaxPol         &  Unsupervised/Non-DL                    &       N.A.        &90            & 25.88  &0.7851\\
\hline
Deblur GAN          &Supervised/DL                    & 17.1         &1                 &  26.65      &   0.7933 \\
\hline
USRNet                 &Unsupervised/DL                      & 6.1          &1              &  24.31   & 0.7396   \\
\hline
XYDeblur                 &Supervised/DL                      & 4.92          &0.729               &   27.18     & 0.8059\\
\hline
Polyblur               &Self-supervised/Non-DL                   & N.A.         &0.001                 & 27.10   &0.7794 \\
\hline
IFFS             &Supervised/DL                      &            19.5    &3.5           &  26.52  &0.8242  \\
\hline
NSSR-DIL              &\begin{tabular}[c]{@{}l@{}}   Image-data\\independent/DL   \end{tabular}             & 0.0028              &0.002             &  25.19 &0.8133\\
\hline
\begin{tabular}[c]{@{}l@{}}$D^3$\\(LCNN)\end{tabular}  &\begin{tabular}[c]{@{}l@{}}   Image-data\\independent/DL   \end{tabular}      &0.0028   & 0.002&  27.01  &\textcolor{blue}{0.8333 }\\
\hline
\begin{tabular}[c]{@{}l@{}}$D^3$\\(DRK)\end{tabular}  &\begin{tabular}[c]{@{}l@{}}   Image-data\\independent/DL   \end{tabular}      &0.00001&0.0005  &  \textcolor{red}{28.02 }   &\textcolor{red}{0.8383}  \\
\hline

\end{tabular}%

\end{table}

\begin{figure}[!t]
    \centering
    \includegraphics[width = \textwidth, height = 7.5cm]{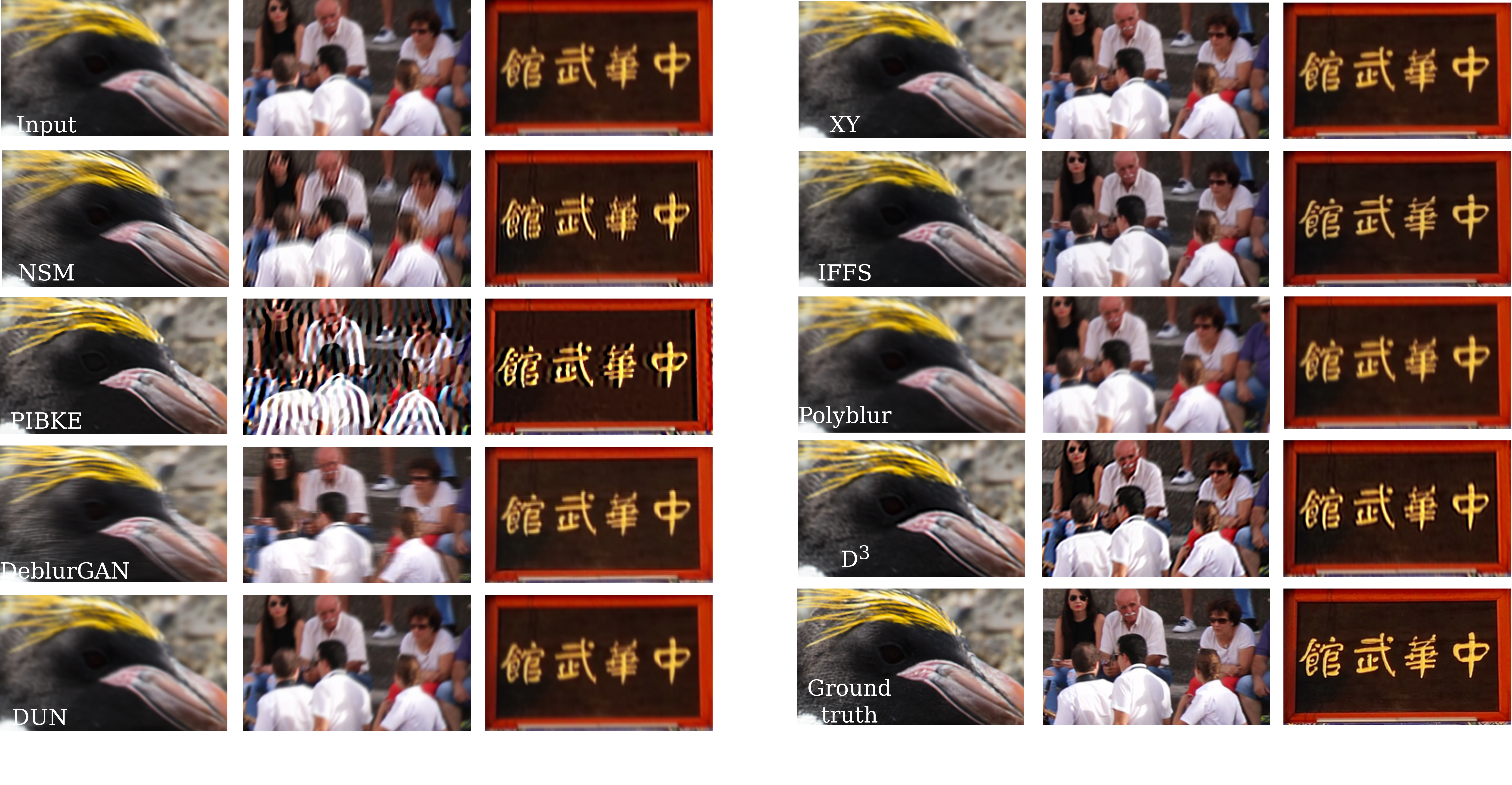}
    \caption{Visual results of different deblur methods.}
    \label{fig:visual-deblur}
\end{figure}
\begin{figure*}
    \centering
    \includegraphics[width = \textwidth]{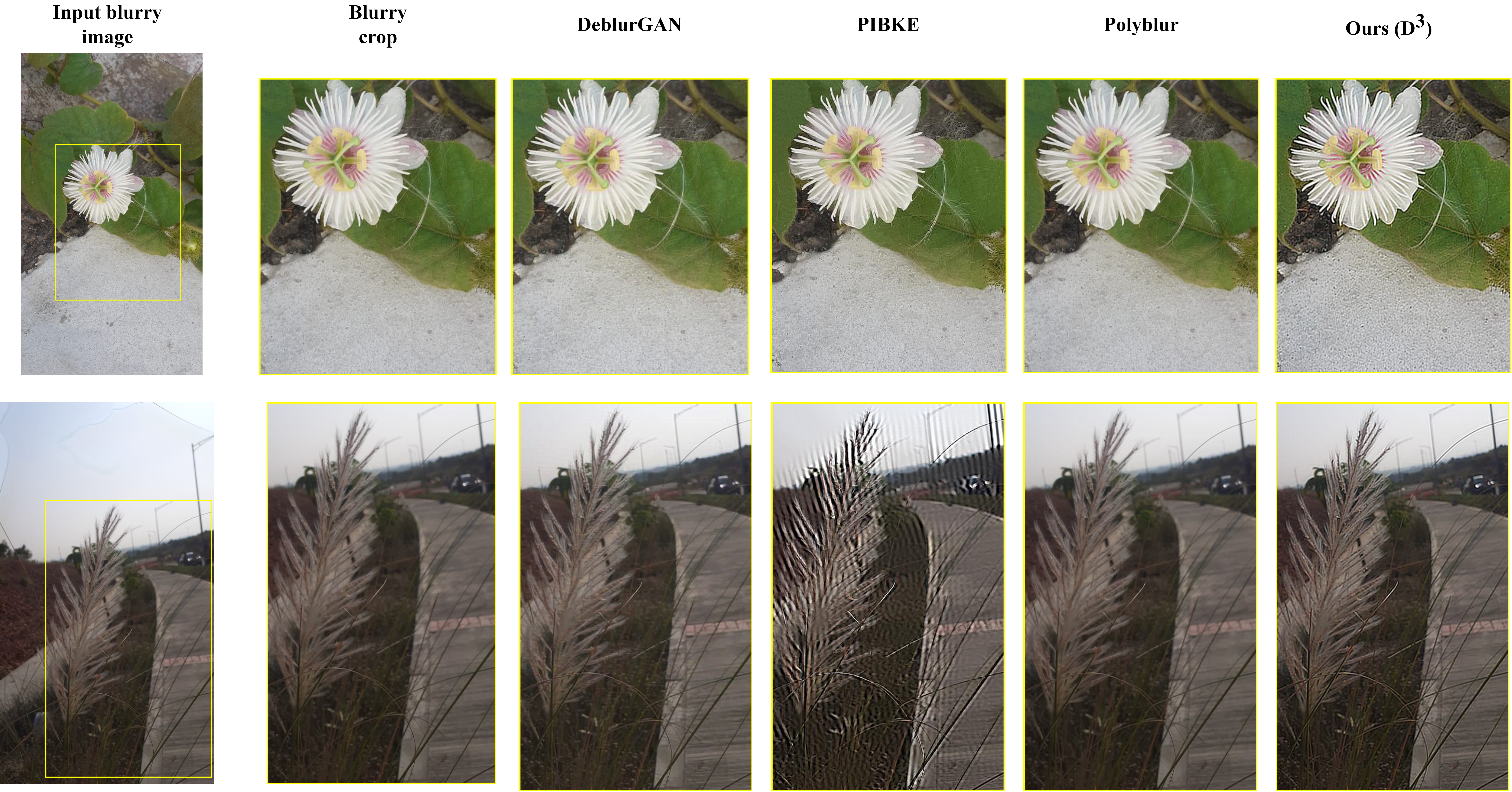}
    \caption{Deblurring performance of various methods on real captured images using smartphone Samsung J7 Prime (13MP, f/1.9)}
    \label{fig:real2}
\end{figure*}

\textbf{Deblur performance comparison:} We generated a  synthetic  blur dataset following the \cite{polyblur}. Random Gaussian kernels of varying shapes, orientations  (i.e. $\sigma_1, \sigma_2 \in U[0.175, 3]$ and rotation angle $\theta \in U[0, \pi]$) and sizes $11 \times 11, 15 \times 15, 19 \times 19, 23 \times 23, 27 \times 27$ were generated. Each set of kernels, varying in size, was applied to the set of 100 sharp images from the DIV2K validation dataset \cite{Timofte_2018_CVPR_Workshops}, resulting in 500 test images. To have a little deviation from the Gaussian distribution, a small amount of multiplicative noise was introduced to the random Gaussian kernels. We compared the $D^3$ model to the above-mentioned SotA works using the simulated blur dataset. To assess the ability to restore the finer details during the deblurring process, we employ PSNR and SSIM for qualitative and quantitative similarity with the corresponding sharp images. The results were tabulated and presented in Table \ref{tab:comparison}. The qualitative results were provided in  Fig. \ref{fig:visual-deblur}.\\
Despite being very lightweight, the proposed $D^3$ demonstrated its effectiveness significantly with standard metrics. The compared SotA methods need supervised learning from images for $K$ estimation and/or deblurring model learning. Thus requires more memory, and time resources at least by order of 100 in parameters and inference time respectively compared to the proposed $D^3$.\\ 
\textbf{Evaluation on real data}: Further, we provide the qualitative comparison of deblurring performance on real captured blurry images provided in Fig. \ref{fig:realresult} and  Fig.\ref{fig:real2}. The qualitative results given in Fig. \ref{fig:visual-deblur}, Fig. \ref{fig:realresult} and  Fig.\ref{fig:real2} illustrate that the proposed model did not produce artifacts in the generated deblurred output images, unlike the compeer, zero-shot non-DL SotA works like \cite{nsm}, \cite{PIBKE}. Besides, the proposed $D^3$ method demonstrates its generalizability by generating sharp images with good dynamic range, without any knowledge of the corresponding blur kernel, has no strict limitations on the blur kernel estimation, and is not restrictive to synthetic kernels in our training dataset.
\subsection{$D^3$ for Image Super-Resolution}
\label{sec:ISR task}
\begin{table}[]
\centering
\caption{Comparison of ISR results on DIV2KRK dataset in terms of   SSIM$\uparrow$,  PSNR$\uparrow$, NIMA$\uparrow$ along with the number of parameters (in Millions) and Inference time (in seconds) of various methods were given below. Here, the \textcolor{red}{red} indicates the best score and the \textcolor{blue}{blue} indicates the second-best score.}
\label{tab:srresults}
\begin{tabular}{lcccccc}
\textbf{Method} &
  \multicolumn{1}{l}{\textbf{\begin{tabular}[c]{@{}l@{}}Scale\\ factor\end{tabular}}} &
  \multicolumn{1}{l}{\textbf{\begin{tabular}[c]{@{}l@{}}Inference\\ time\\ (in seconds)\end{tabular}}} &
  \multicolumn{1}{l}{\textbf{\begin{tabular}[c]{@{}l@{}}\#Parameters\\ (in Millions)\end{tabular}}} &
  \multicolumn{1}{l}{\textbf{SSIM}} &
  \multicolumn{1}{l}{\textbf{PSNR}} &
  \multicolumn{1}{l}{\textbf{NIMA}} \\ \hline
\multirow{2}{*}{Bicubic} &
  X2 &
  \multirow{2}{*}{N.A.} &
  \multirow{2}{*}{N.A.} &
  0.7846 &
  27.24 &
  4.127 \\   
 &
  X4 &
   &
   &
  0.6478 &
  23.89 &
  4.147 \\  \hline
\multirow{2}{*}{ZSSR} &
  X2 &
  \multirow{2}{*}{0.29} &
  \multirow{2}{*}{=10} &
  0.7925 &
  27.51 &
  4.111 \\  
 &
  X4 &
   &
   &
  0.6550 &
  24.05 &
  4.156 \\  \hline
\multirow{2}{*}{\begin{tabular}[c]{@{}l@{}}KernelGAN\\ +ZSSR\end{tabular}} &
  X2 &
  \multirow{2}{*}{0.151+0.29} &
  \multirow{2}{*}{\textgreater{}=13} &
  0.8379 &
 \textcolor{blue}{  28.24 }&
  4.071 \\
 &
  X4 &
   &
   &
  0.6799 &
 \textcolor{blue}{  24.76 }&
  4.089 \\ \hline
\multirow{2}{*}{DBPI} &
  X2 &
  \multirow{2}{*}{0.5} &
  \multirow{2}{*}{\textgreater{}=1} &
  0.8684 &
 \textcolor{red}{ 30.77} &
  4.049 \\
 &
  X4 &
   &
   &
  0.7368 &
  \textcolor{red}{26.86} &
  4.146 \\ \hline
\multirow{2}{*}{DualSR} &
  X2 &
  \multirow{2}{*}{0.44} &
  \multirow{2}{*}{\textgreater{}=3} &
  0.8250 &
  25.78 &
 \textcolor{blue}{ 5.005} \\
 &
  X4 &
   &
   &
  - &
  - &
  - \\ \hline
\multirow{2}{*}{NSSR-DIL} &
  X2 &
  \multirow{2}{*}{0.0028} &
  \multirow{2}{*}{0.002} &
  0.8644 &
  26.02 &
  4.161 \\
 &
  X4 &
   &
   &
 \textcolor{red}{ 0.7926} &
  23.58 &
  4.170 \\ \hline
\multirow{2}{*}{\begin{tabular}[c]{@{}l@{}}$D^3$\\ (LCNN)\end{tabular}} &
  X2 &
  \multirow{2}{*}{0.0028} &
  \multirow{2}{*}{0.002} &
\textcolor{blue}{  0.8980} &
  27.53 &
{   4.976} \\
 &
  X4 &
   &
   &
  0.7331 &
  23.63 &
  \textcolor{red}{4.904} \\ \hline
\multirow{2}{*}{\begin{tabular}[c]{@{}l@{}}$D^3$\\ (DRK)\end{tabular}} &
  X2 &
  \multirow{2}{*}{0.00001} &
  \multirow{2}{*}{0.0005} &
 \textcolor{red}{ 0.9038 }&
  27.27 &
 \textcolor{red}{ 5.006 }\\
 &
  X4 &
   &
   &
 \textcolor{blue}{  0.7487 }&
  23.93 &
 \textcolor{blue}{  4.903} \\
  \hline
\end{tabular}%
\end{table}

\begin{figure}[!t]
    \centering
    \includegraphics[width = \textwidth, height = 7.5 cm]{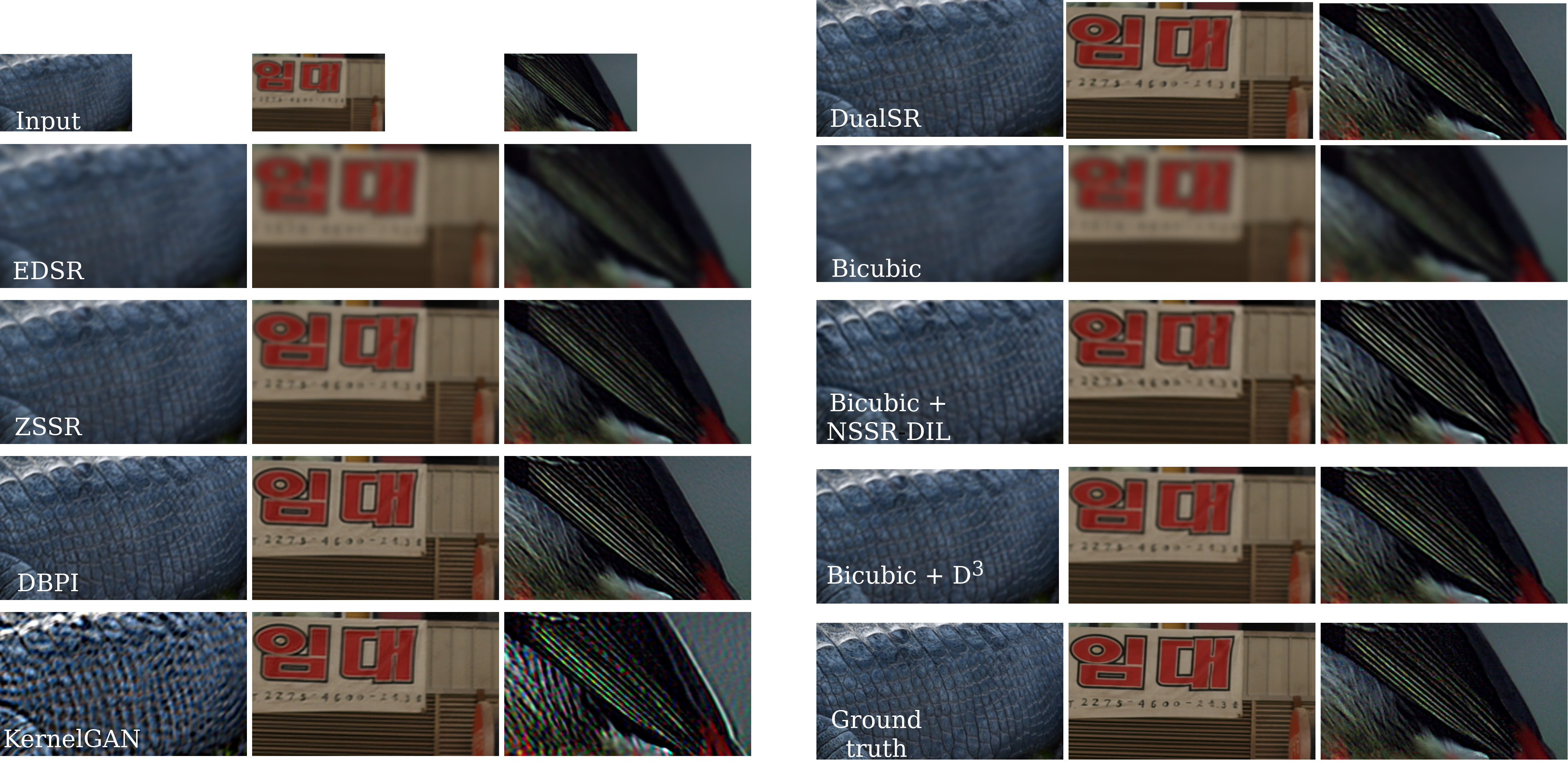}
    \caption{Visual results of different ISR methods.}
    \label{fig:visual-sr}
\end{figure}
Image Super Resolution (ISR) is a well-established low-level vision task whose objective is to generate a High-Resolution (HR) image from the given corresponding LR observation(s). We propose to extend the proposed model to the ISR task. The $D^3$ model can be directly employed at the inference stage, as the trained L-CNN acts as a pre-upsampled network that operates on traditional upsampled (Bicubic) input blur images of any desired scale factor to generate sharp images with fine details. The same trained L-CNN for the deblurring task was employed to perform the ISR task. We also note that this DRK alone could also be employed for the ISR task in place of L-CNN on a pre-upsampled image and operates the same as L-CNN. In these experiments, no kernel estimation stage is required. The super-resolution ability of our approach was evaluated on the benchmark dataset DIV2KRK \cite{kernelgan} for sf 2, and sf 4 using PSNR, SSIM, and NIMA metrics. The ISR results were presented in Table \ref{tab:srresults}. For a fair comparison, we considered works like KernelGAN \cite{kernelgan}, ZSSR \cite{zssr}, DBPI \cite{dbpi}, and DualSR \cite{emad2021dualsr} which consider single input LR image only for modeling the ISR task as in our experiments. From the quantitative results, it was observed that the $D^3$ model delivers competitive performance very close to the SotA ISR works despite being absolutely independent of LR and HR image space. The sample visual results for sf 2 were provided in  Fig. \ref{fig:visual-sr}. The computational complexity in terms of parameters and super-resolution time (inference time) in minutes, of the $D^3$ model was compared to SotA methods and is given in Table \ref{tab:srresults}. The potential of the proposed $D^3$ model is observed from the given Table \ref{tab:srresults}. Here, the $D^3$ requires approximately 100 times fewer memory resources even when compared with the least among the SotA methods i.e. ZSSR, where the DRK requires by order 10 times further less than $D^3$. Besides, the inference time required for the $D^3$ and the DRK is reduced by order of 10 and 100 respectively compared to existent zero-shot ISR methods.\\ 
Notably, while both our current proposed method and previous work (i.e. NSSR-DIL) utilize the L-CNN model, the $D^3$ framework exhibits marked gains in deblurring and ISR performance compared to our earlier approach, primarily attributable to the updated regularization term.
\subsection{Ablation study}
In this section, the significance of the proposed regularization term ($R$) (refer Eq. \ref{eq5}), and the deblurring performance of the $D^3$ method on various blur levels along with varied kernel sizes and various sizes of the RKG dataset were discussed. Further, we present the comparison of the deblurring performance of our proposed method with the traditional deconvolution filter i.e. Wiener filter.
\subsubsection{Regularization term (R)}
The regularization term ($R$) (refer Eq. \ref{eq5}) is proposed to have a faithful deconvolution model with reliable deblurring performance. The influence of each entity in the proposed $R$ is quantified and presented in Table \ref{tab:R}. It was observed that the presence of $L_{ConvArea}$ in $R$ has shown a greater impact on the performance of the proposed method. The terms $L_{PhLoss} ~\&~ L_{MagLoss}$ influenced the perceptual quality of the images. The term $L_{ConvArea}$ together with the $L_{PhLoss} ~\&~ L_{MagLoss}$ achieve a reliable performance.\\
\begin{table}[!h]

\centering
\footnotesize
\caption{The effect of proposed regularization terms in R on the $D^3$ model's performance with the synthetic Gaussian blur dataset.}
\label{tab:R}
\begin{tabular}{ccc}
\textbf{Loss (L) }  & \textbf{SSIM$\uparrow$} & \textbf{PSNR$\uparrow$} \\ \hline \\ 
$ \lVert K \ast K^{-1} - ~\delta \rVert^2_2 $  & 0.0224 & 6.59 \\ \hline \\ 
$\lVert K \ast K^{-1} - ~\delta \rVert^2_2 +   \lambda_1 \times L_{ConvArea}$& 0.5455 & 21.07\\ \hline \\ 
$ \lVert K \ast K^{-1} - ~\delta \rVert^2_2 + \lambda_2 \times L_{PhLoss}$ & 0.2137 & 8.12\\ \hline \\ 
$ \lVert K \ast K^{-1} - ~\delta \rVert^2_2 + \lambda_3 \times L_{MagLoss}$ & 0.2030 & 17.69\\ \hline \\  
$ \lVert K \ast K^{-1} - ~\delta \rVert^2_2 + \lambda_2 \times L_{PhLoss}  + \lambda_3 \times L_{MagLoss}$& 0.5644 & 17.21 \\  \hline  \\ 
$ \lVert K \ast K^{-1} - ~\delta \rVert^2_2 +   \lambda_1 \times L_{ConvArea} + \lambda_2 \times L_{PhLoss}  $& 0.7419 & 25.04  \\  \hline \\ 
$ \lVert K \ast K^{-1} - ~\delta \rVert^2_2 +   \lambda_1 \times L_{ConvArea} + \lambda_2 \times L_{PhLoss}  + \lambda_3 \times L_{MagLoss}$& 0.8333 & 27.01 \\ \hline
\end{tabular}%
\end{table}

\subsubsection{Evaluation on various blur levels:}
\label{extendedblur}
We evaluated the deblurring performance of the proposed $D^3$ model on various blur levels, whose blur kernels are disjoint with our training dataset. We synthetically generated a blur dataset following similar techniques used to generate a blur test dataset as discussed in Sec. 4 with different sets of variances $\sigma_1, \sigma_2 \in U[0.175, 3], U[3, 6], ~\&~ U[6, 9]$
and using different kernel sizes i.e. $11 \times 11, 15 \times 15, 19 \times 19 ~ and ~ 21 \times 21$. 
The results were outlined in Table \ref{tab:beyond}.  $D^3$ model trained on $11 \times 11$ size kernels, demonstrates robustness even when tested against a broad spectrum of blurs, encompassing varied kernel sizes such as $11 \times 11, 15 \times 15, 19 \times 19 ~ and ~ 21 \times 21$. The experimental findings showcase the overall effectiveness and satisfactory deblurring capabilities of the $D^3$ model.

\begin{table*}[!h]
\centering
\caption{Deblurring performance of the proposed $D^3$ model on various blur levels using PSNR/SSIM metrics.}
\label{tab:beyond}
\resizebox{0.8\textwidth}{!}{%
\begin{tabular}{|cc|cccc|}
\hline
\multicolumn{2}{|c|}{\multirow{2}{*}{}} & \multicolumn{4}{|c|}{Degradation kernel $K$ size} \\  
\multicolumn{2}{|c|}{} & \multicolumn{1}{|c|}{$11 \times 11$} & \multicolumn{1}{|c|}{$15 \times 15$} & \multicolumn{1}{|c|}{$19 \times 19$} & $21 \times 21$ \\ \hline
 \multicolumn{1}{|c|}{\multirow{3}{*}{Variance levels}} & $\sigma_1, \sigma_2 \in U[0.1, 3] $  & \multicolumn{1}{|c|}{26.67/0.8436} & \multicolumn{1}{|c|}{26.53/0.8451} & \multicolumn{1}{|c|}{26.53/0.8441} & 26.61/0.8440 \\ 
\multicolumn{1}{|c|}{} & $\sigma_1, \sigma_2 \in U[3,6] $ & \multicolumn{1}{|c|}{25.92/0.7726} & \multicolumn{1}{|c|}{25.89/0.7727} & \multicolumn{1}{|c|}{25.79/0.7697} & 25.84/0.7720 \\ 
\multicolumn{1}{|c|}{} & $\sigma_1, \sigma_2 \in U[6,9] $ & \multicolumn{1}{|c|}{25.02/0.7220} & \multicolumn{1}{|c|}{24.74/0.7119} & \multicolumn{1}{|c|}{24.72/0.7124} & 24.68/0.7102 \\ \hline
\end{tabular}%
 }
\end{table*}
\subsubsection{Performance of $D^3$ with varied sizes of RKG dataset}
\begin{figure}[!h]
    \centering
    \includegraphics[width = 0.95\textwidth, height = 4.75cm]{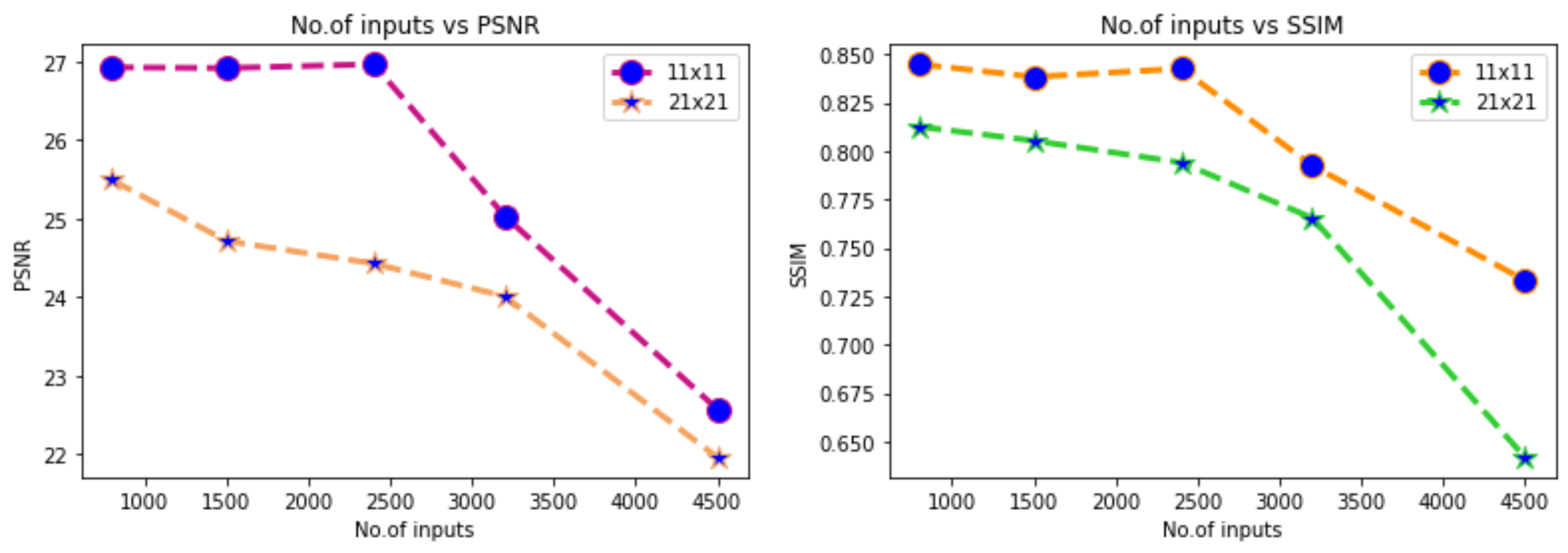 }
    \caption{The performance comparison of the $D^3$ model with sizes of RKG dataset.}
    \label{fig:samples}
\end{figure}

We studied the performance of the $D^3$ model trained on varied sizes of the RKG dataset to empirically find the size of the RKG dataset to train the proposed $D^3$ model. In our study, we generated five different datasets with the number of samples $800, 1600, 2400, 3200, 4000$, and $4800$ and named as $RKG_n$, where $n = 1, 2, 3, 4, 5$, respectively. These five datasets i.e. $RKG_n$ were generated for kernel sizes $11 \times 11 ~\&~ 21 \times 21$, following the steps discussed in the case of RKG dataset generation. The L-CNN was trained independently on each set and evaluated on the synthetic Gaussian blur dataset using PSNR and SSIM metrics. The visual plots depicting the L-CNN performance comparison for $RKG_n$ vs evaluation metrics i.e. PSNR, and SSIM are presented in Fig. \ref{fig:samples} (a) and Fig. \ref{fig:samples} (b) respectively. 
It is observed that there is a linear improvement in these metrics as the number of samples was increased initially and later fell off the clip. Therefore we considered 2400 training samples i.e. $RKG_3$ dataset, with kernel size $11 \times 11$, in all our experiments.

\subsubsection{Deblurring performance comparison of the proposed method with Wiener filter.} In our experiments, we compared the deblurring performance of our proposed learnable deconvolution kernel with the traditional inverse filter, i.e. the Wiener filter. We observed that the performance of the Wiener filter is not robust across the dataset and shows $>$50\% poor values in evaluation metrics, PSNR, and SSIM. We also trained a Single Layer and Single Channel (SLSC) linear CNN, to compute the generalized inverse kernel for the RKG dataset using the DIL objective and the proposed regularization term. The results were tabulated in Table \ref{WienerFilter}. In contrast to the performance of traditional Wiener filter's and SLSC networks, the proposed method can handle a wide range of degradations across the dataset and generate deblurred images with fine details. 
\begin{table}[!h]
\centering
\small
\caption{Comparison of deblurring performance of $D^3$ and Wiener filter on the generated synthetic blur dataset, discussed in Section 4.3.}
\label{WienerFilter}
\begin{tabular}{|c|c|c|}
\hline
\textbf{Method}           &   \textbf{PSNR} & \textbf{SSIM} \\ \hline
Wiener Filter             & 10.49     & 0.1752       \\ \hline 
$D^3$ (L-CNN-SLSC)   &   19.28 & 0.5248 \\ \hline
$D^3$ (L-CNN)                  & 27.01    & 0.8333     \\ \hline
$D^3$ (DRK)                 & 28.02      & 0.8383        \\ \hline
\end{tabular}%
\end{table}

\section{Conclusion}
In this work, our goal is to design a deblur model to restore sharp images from blurry inputs for practical applications. Therefore, we proposed $D^3$, a computationally efficient deblur model and the first DL-based image data-independent model that involves a restoration process rather than learning a mapping from distorted to sharp image distribution. Our proposed deblur framework can be directly extended to the ISR task as well. The superior performance of the proposed method can be attributed to two main reasons: the DIL objective and the RKG dataset. This work demonstrates tremendous hope for improving the deblur and ISR capability without the need for image datasets (i.e. both supervised and unsupervised ISR datasets). Our $D^3$ method paves the path to have a deeper look into the learning and understanding of the inverse degradation kernel since the explicit representation i.e. DRK, in a matrix form is made possible. Additionally, we have demonstrated the deblurring task of replacing the deep model with a simple convolution operation with the computed DRK, which doesn't require high-performance computing resources. Despite being a very lightweight deblurring model, the proposed method's performance is superior to the SotA works. The experimental results and computational performance comparisons with the SotA indicate that the proposed $D^3$ and the DRK are remarkably suitable for practical embedded applications of deblurring and ISR tasks.

\bibliography{sn-bibliography.bib}

\end{document}